\documentclass{article}

\usepackage{microtype}
\usepackage{graphicx}
\usepackage[export]{adjustbox}  %
\usepackage{subcaption}
\usepackage{booktabs} %

\usepackage{hyperref}

\usepackage[preprint]{icml2026}

\usepackage{amsmath}
\usepackage{amssymb}
\usepackage{mathtools}

\usepackage{xcolor}
\usepackage{enumitem}
\usepackage{tikz}
\usetikzlibrary{positioning}
\usepackage{multirow}

\newenvironment{denseitemize}{
	\begin{itemize}[topsep=2pt, partopsep=0pt, leftmargin=1.5em]
		\setlength{\parskip}{0pt}
		\setlength{\parsep}{0pt}
	}{\end{itemize}}

\newenvironment{denseenum}{
	\begin{enumerate}[topsep=2pt, partopsep=0pt, leftmargin=1.5em]
		\setlength{\itemsep}{2pt}
		\setlength{\parskip}{0pt}
		\setlength{\parsep}{0pt}
	}{\end{enumerate}}

\usepackage{tcolorbox}
\newcounter{observation}
\newtcolorbox{takeaway}{
  colback=gray!10,
  colframe=gray!50,
  boxrule=0.5pt,
  arc=2pt,
  left=6pt,
  right=6pt,
  top=4pt,
  bottom=4pt,
  before upper={\refstepcounter{observation}\textbf{Observation \theobservation.}\ },
}

\icmltitlerunning{Where Do the Joules Go? Diagnosing Inference Energy Consumption}
 
\begin{document}

\twocolumn[
  \icmltitle{Where Do the Joules Go? Diagnosing Inference Energy Consumption}

  \icmlsetsymbol{equal}{*}

  \begin{icmlauthorlist}
    \icmlauthor{Jae-Won Chung}{umich}
    \icmlauthor{Ruofan Wu}{umich}
    \icmlauthor{Jeff J. Ma}{umich}
    \icmlauthor{Mosharaf Chowdhury}{umich}
  \end{icmlauthorlist}

  \icmlaffiliation{umich}{University of Michigan \& The ML.ENERGY Initiative}

  \icmlcorrespondingauthor{Jae-Won Chung}{jwnchung@umich.edu}

  \icmlkeywords{energy, inference, LLM, diffusion, GPU, benchmark}

  \vskip 0.3in
]

\printAffiliationsAndNotice{}

\begin{abstract}
Energy is now a critical ML computing resource.
While measuring energy consumption and observing trends is a valuable first step, accurately understanding and diagnosing \emph{why} those differences occur is crucial for optimization.
To that end, we begin by presenting a large-scale measurement study of inference time and energy across the generative AI landscape with 46 models, 7 tasks, and 1,858 different configurations on NVIDIA H100 and B200 GPUs.
Our empirical findings span order-of-magnitude variations: LLM task type can lead to 25$\times$ energy differences, video generation sometimes consumes more than 100$\times$ the energy of images, and GPU utilization differences can result in 3--5$\times$ energy differences.
Based on our observations, we present a framework for \emph{reasoning} about the underlying mechanisms that govern time and energy consumption.
The essence is that time and energy are determined by \emph{latent} metrics like memory and utilization, which are in turn affected by various factors across the algorithm, software, and hardware layers.
Our framework also extends directly to throughput per watt, a critical metric for power-constrained datacenters.

\end{abstract}

\section{Introduction}\label{sec:intro}

As AI infrastructure scales to meet rapid growth in AI compute demand, energy consumption is becoming a critical bottleneck~\cite{cbre2023,cbre2024,cbre2025,mckinsey2023,mckinsey2024,mckinsdy2025,zuckerberg-ai-energy,electricity-iea25,bloombergnef25}.
This is primarily due to a mismatch between energy demand and supply---building new power sources to supply energy takes years of planning, approval, and construction~\cite{eia-utility-data,ai-grid-impact-arxiv25}, whereas AI compute demand skyrockets relentlessly.

Yet, energy has not been a mainstream metric for the ML community, which has led to a lack of tools to measure and understand energy consumption.
Recently, works like the ML.ENERGY Benchmark~\cite{mlenergy-neurips25} helped fill this gap with tools that provide energy measurements in environments that are representative of real-world deployments.
However, while measuring energy is necessary, it is not sufficient; without understanding the underlying mechanisms that govern the measured end metrics, we cannot reason about why one configuration consumes more time or energy than another, nor can we guide ourselves through the optimization space.

The goal of this paper is to provide a framework for understanding and reasoning about inference energy consumption in generative AI workloads.
To do so, we begin with a large-scale empirical study of inference time and energy consumption across up-to-date models, tasks, software systems, and hardware---spanning 46 models across 7 tasks, producing 1,858 configurations on NVIDIA H100 and B200 GPUs (\S\ref{sec:energy-by-arch}).
We observe order-of-magnitude variations driven within and across models and tasks:
LLM task type can lead to 25$\times$ energy differences,
video generation sometimes consumes more than 100$\times$ the energy of images,
and GPU utilization differences can result in 3--5$\times$ energy gaps.

Given this, we dive deeper into the specific factors that affect energy consumption with controlled comparisons (\S\ref{sec:deeper-dive}), including knobs at the model-level and system-level.
We uncover several puzzling observations: lower precision is not always faster or more energy efficient, and increasing the number of GPUs can consume \emph{less} total energy by unlocking larger memory capacity.
These counterintuitive findings provide clues into the underlying mechanisms that govern time and energy.

Finally, based on our analysis, we develop a framework for reasoning about energy consumption (\S\ref{sec:reasoning}).
Energy consumption is governed by various \emph{latent} factors that are not directly observable from end metrics alone, like memory availability and usage, hardware utilization, and application constraints.
Our framework not only is useful for wall-clock time and energy consumption measurements, but also extends to reasoning about the service capacity of AI datacenters that are constrained by power budgets.

\section{Methodology}\label{sec:methodology}

Our goal is to measure time and energy consumption in a production-representative manner.
We focus on GPU energy, as AI accelerators account for 50--70\% of datacenter power/energy~\cite{polca-asplos24,gemini-energy-arxiv25}.

Our methodology follows that of the ML.ENERGY Benchmark~\cite{mlenergy-neurips25}, using Zeus~\cite{zeus-nsdi23,zeus-github} for energy measurement, but we have upgraded to incorporate up-to-date LLMs including Qwen 3~\cite{qwen3-arxiv25}, DeepSeek R1~\cite{deepseek-r1-arxiv25}, GPT OSS~\cite{gpt-oss-arxiv25}, and the latest diffusion models.
We use a production-grade serving stack: vLLM 0.11.1~\cite{pagedattention-sosp23} for LLMs and MLLMs and xDiT~\cite{xdit-arxiv24} 0.4.5 for diffusion models, on NVIDIA H100 and B200 GPU nodes with NVSwitch support.

For LLMs, we identify the steady state when batch size\footnote{For vLLM, the max batch size (\texttt{max\_num\_seqs}) configuration for the server. For xDiT, standard inference batch size.} saturates and compute energy per token by dividing steady-state total energy consumption by the number of tokens generated during steady state.
For diffusion models, which are batched as a whole, per-request energy is the total energy consumption of the batch divided by batch size.
Batch sizes and GPU counts are swept for each model with BF16 as the default precision.
When a native FP8 weight LLM or MLLM is available, we include it as a separate model.
See Appendix~\ref{apdx:tasks-and-models} for the full task, dataset, and model list.

\section{Energy by Architecture}\label{sec:energy-by-arch}

In this section, we broadly compare the energy consumption of different model architectures: LLMs (\S\ref{sec:llm}), multimodal LLMs (\S\ref{sec:mllm}), and diffusion models (\S\ref{sec:diffusion}).

\subsection{Large Language Models}\label{sec:llm}

\paragraph{Task type heavily influences output length.}
LLM time and energy consumption is dominated by the decoding (token generation) phase~\cite{mlenergy-neurips25}.
Different tasks naturally produce different distributions of output lengths.
This is particularly pronounced between two LLM tasks in our benchmark: Problem Solving (reasoning on) and Text Conversation (reasoning off).

\begin{figure}[t]
  \centering
  \begin{subfigure}[b]{0.325\columnwidth}
    \centering
    \includegraphics[width=\columnwidth]{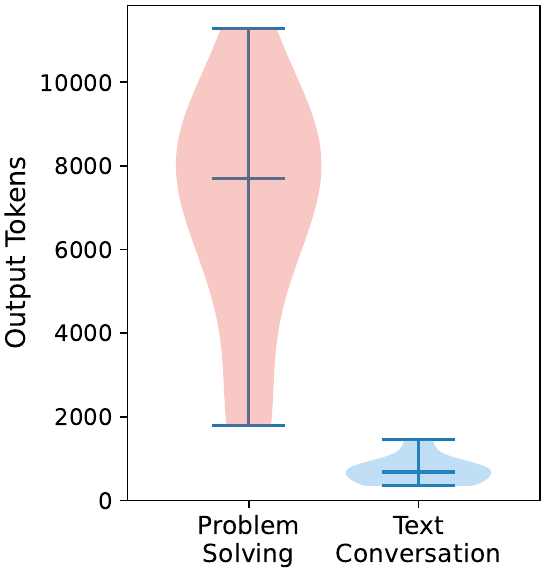}
    \caption{Output length}\label{fig:llm-energy-a}
  \end{subfigure}
  \begin{subfigure}[b]{0.30\columnwidth}
    \centering
    \includegraphics[width=\columnwidth]{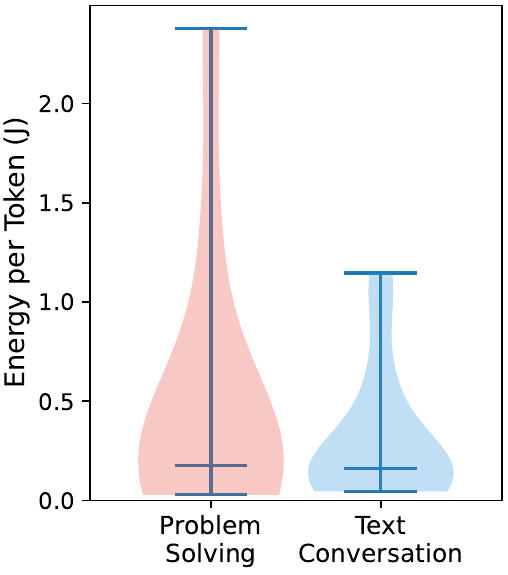}
    \caption{Energy/token}\label{fig:llm-energy-b}
  \end{subfigure}
  \begin{subfigure}[b]{0.32\columnwidth}
    \centering
    \includegraphics[width=\columnwidth]{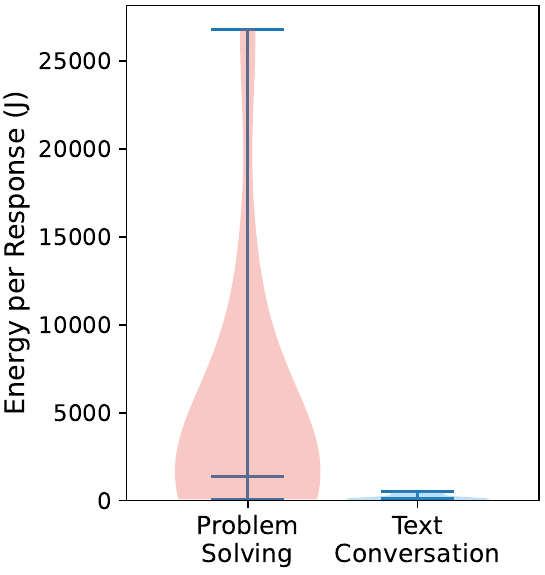}
    \caption{Energy/response}\label{fig:llm-energy-c}
  \end{subfigure}
  \caption{Distribution of (a) number of output tokens, (b) energy per token, and (c) energy per response across all models on B200 GPUs using their respective minimum-energy configurations.}\label{fig:llm-energy}
\end{figure}

Figure~\ref{fig:llm-energy} compares B200 minimum-energy configurations,\footnote{The configuration (e.g., batch size, number of GPUs) that achieves the lowest energy consumption for the model.} focusing on model and task differences without hardware utilization confounds.
Problem Solving generates on average 10$\times$ more output tokens than Text Conversation (mean 6,988 vs.\ 717).
Additionally, longer output sequences stress memory capacity and prevent larger batch sizes, increasing energy per token due to lower GPU utilization.
Since energy per response is energy per token multiplied by the number of output tokens, these two factors multiply, resulting in Problem Solving consuming on average 25$\times$ more energy per response than Text Conversation (mean 4,625~J vs.\ 184~J).

\paragraph{Case study.}
Qwen 3 32B supports both reasoning and non-reasoning, enabling direct comparison on the same model.

\begin{table}[t]
  \footnotesize
  \centering
  \caption{Qwen 3 32B comparison across tasks on 1$\times$ B200.}
  \label{tab:qwen3-comparison}
  \begin{tabular}{@{}lrrr@{}}
    \toprule
    \textbf{Metric} & \shortstack[r]{\textbf{Text}\\\textbf{Conv.}} & \shortstack[r]{\textbf{Problem}\\\textbf{Solving}} & \textbf{Ratio} \\
    \midrule
    Max batch size (BS) & 512 & 128 & $0.25\times$ \\
    Mean output tokens & 627 & 7,035 & 11$\times$ \\
    Energy/tok @ BS 128 & 0.209 J & 0.312 J & 1.5$\times$ \\
    Energy/tok @ max BS & 0.151 J & 0.312 J & 2.1$\times$ \\
    Energy/response & 95 J & 2,192 J & 23$\times$ \\
    \bottomrule
  \end{tabular}
\end{table}

As shown in Table~\ref{tab:qwen3-comparison}, longer output sequences in Problem Solving increase the amount of KV cache memory usage per response, preventing larger batch sizes.
Therefore, when we compare energy per token at each task's maximum batch size, Problem Solving is 2.1$\times$ higher.
Even at the same batch size (128), longer sequences consume more energy per token due to higher memory footprint.
Finally, combining longer outputs and higher energy per token results in 23$\times$ energy per response for Problem Solving for this model.

\begin{takeaway}
Task type heavily influences energy consumption.
Notably, Problem Solving (reasoning) uses on average 25$\times$ more energy per response than Text Conversation.
This comes from 10$\times$ more output tokens combined with higher energy per token due to memory pressure limiting batch size.
\end{takeaway}

\subsection{Multimodal LLMs}\label{sec:mllm}

Multimodal LLMs (MLLMs) take images and/or videos alongside text as input and generate text responses.

\begin{figure}[t]
  \centering
  \begin{subfigure}[b]{0.44\columnwidth}
    \centering
    \includegraphics[width=0.95\columnwidth]{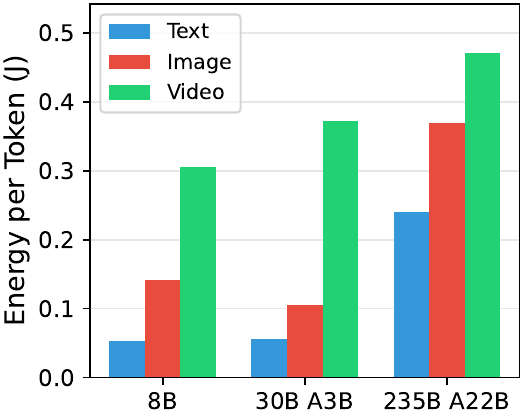}
    \caption{Energy/token by modality}
    \label{fig:mllm-energy-a}
  \end{subfigure}
  \hfill
  \begin{subfigure}[b]{0.51\columnwidth}
    \centering
    \includegraphics[width=0.99\columnwidth]{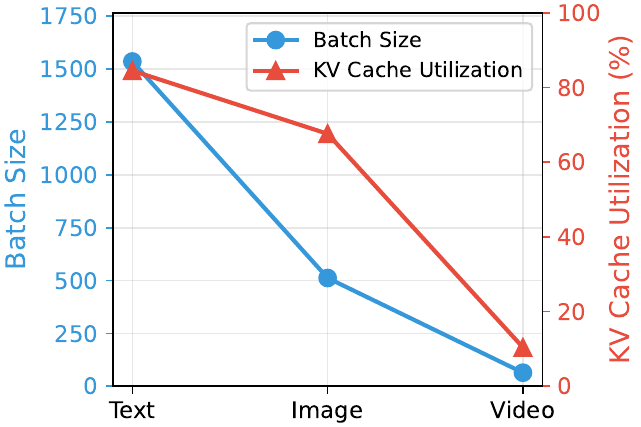}
    \caption{Qwen 3 VL 8B, 1$\times$ B200}
    \label{fig:mllm-energy-b}
  \end{subfigure}
  \caption{
    (a) shows three models from the Qwen 3 family across three modalities (minimum-energy configurations), and (b) shows how modality affects batch size and KV cache utilization for the 8B model, showing why energy per token increases.
    Text modality uses the text-only model (e.g., Qwen 3 8B), whereas Image and Video use the vision--language variant (e.g., Qwen 3 VL 8B).
  }\label{fig:mllm-energy}
\end{figure}

\newpage
\paragraph{Multimodality can increase energy.}
The implications of multimodal inputs are threefold:
\begin{denseenum}
  \item Models run their modality encoder to convert inputs into multimodal tokens, which increases computation and memory operations and therefore energy consumption.
  \item In GPU memory-constrained scenarios, the modality encoder and the increase in input length increase memory usage, which can limit batch size.
  \item Multimodal inputs need to be preprocessed first on the CPU-side (e.g., converting raw image/video into tiles of pixels), which can take non-negligible time and become a bottleneck that further limits batch size.
\end{denseenum}

Indeed, as shown by Figure~\ref{fig:mllm-energy-a}, when we compare minimum-energy configurations for different modalities, text + image inputs use 1.1--5.2$\times$ the energy per token of text, while text + video inputs use 1.3--15.0$\times$.

\paragraph{Case study.}
We compare Qwen 3 8B on Text Conversation with Qwen 3 VL 8B on Image Chat and Video Chat tasks.
For this smaller 8B model, the overheads of vision encoders and CPU-side preprocessing limit batch size significantly and underutilize the GPU, as shown in Figure~\ref{fig:mllm-energy-b}.
In particular, video inputs typically get converted to more vision tokens and are more expensive to preprocess on the CPU side, as shown by the much smaller batch size and higher energy per token.
The drop in KV cache utilization as vision preprocessing overhead grows confirms that GPU memory was not the limiting factor---there was spare capacity for more tokens---but CPU-side vision preprocessing became a severe bottleneck that limited batch size.

All in all, this is a case where GPU energy consumption is not just about the GPU; the entire system and the location of bottlenecks matter.
If CPU-side processing speed remains unchanged and only the GPU is upgraded, the GPU will only be more underutilized.
In subsequent analyses, we do not include MLLMs because CPU-side bottlenecks make it difficult to isolate factors that impact GPU energy.

\begin{takeaway}
Multimodal inputs cost 1.1--5.2$\times$ (image) and 1.3--15.0$\times$ (video) the energy per token of text.
CPU-side vision preprocessing can be a bottleneck that reduces batch size and increases energy per token.
\end{takeaway}

\subsection{Diffusion Models}\label{sec:diffusion}

We benchmarked diffusion models that generate images and videos from user text prompts.
Diffusion is where model size is not the best predictor of energy consumption due to multiple \emph{runtime} factors: number of inference (denoising) steps, output resolution, and number of frames (for video).

\begin{figure}[t]
  \centering
  \begin{subfigure}[b]{0.48\columnwidth}
    \centering
    \includegraphics[width=\columnwidth]{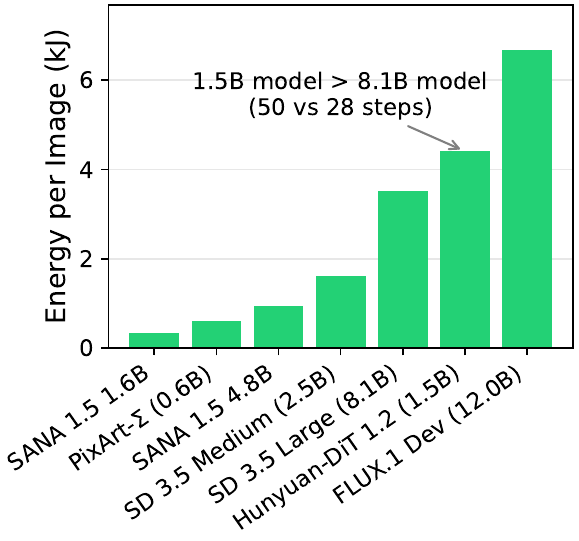}
    \caption{Text to Image}
    \label{fig:diffusion-energy-a}
  \end{subfigure}
  \hfill
  \begin{subfigure}[b]{0.48\columnwidth}
    \centering
    \includegraphics[width=\columnwidth]{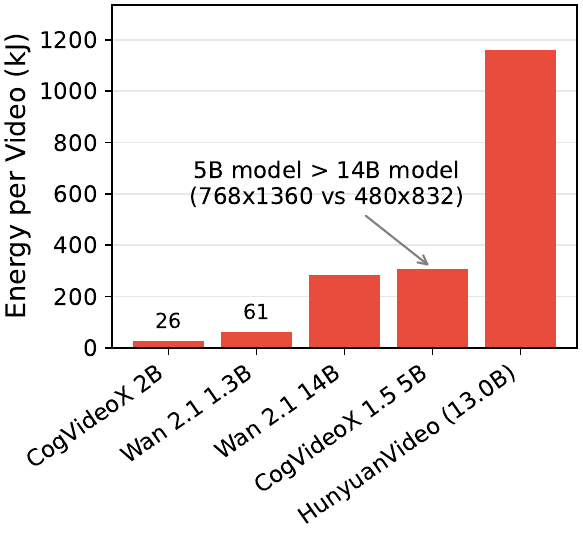}
    \caption{Text to Video}
    \label{fig:diffusion-energy-b}
  \end{subfigure}
  \caption{Energy per image/video for diffusion models (minimum-energy configuration on B200). SD is short for Stable Diffusion.}
  \label{fig:diffusion-energy}
\end{figure}

\paragraph{Text to Image varies 20$\times$ across models.}
Models range from 0.6B to 12B parameters with 20--50 denoising steps (Figure~\ref{fig:diffusion-energy}).
Notably, Hunyuan-DiT 1.2 (1.5B) consumes more energy than SD 3.5 Large (8.1B) despite fewer parameters, largely due to running 50 vs.\ 28 denoising steps.

\paragraph{Text to Video can be very energy intensive.}
Generating a single video consumes 26 kJ to 1.16 MJ---one to two orders of magnitude more than images.
CogVideoX 1.5 5B uses more energy than Wan 2.1 14B despite being smaller, largely because it generates at higher resolution (768$\times$1360 vs.\ 480$\times$832).
HunyuanVideo reaches 1.16 MJ because it generates 129 frames at 720p, resulting in 4$\times$ higher energy than Wan 2.1 14B (13B vs.\ 14B).

\looseness=-1
We used default runtime parameters (denoising steps, resolution, frames) for all models.
Many of these parameters are \emph{controllable by users}, enabling navigation of the time--energy--quality tradeoff space.
The ML.ENERGY Benchmark~\cite{mlenergy-neurips25} explored this for diffusion models.

\begin{takeaway}
Diffusion model energy depends on more than model size: number of denoising steps, output resolution, and frame count matter as much or more.
Video generation can consume one to two orders of magnitude more energy than image generation.
\end{takeaway}

\section{Deeper Dive into Energy}\label{sec:deeper-dive}

In this section, we measure and observe how different factors affect energy consumption.

\subsection{Batch Size}\label{sec:batch-size}

\begin{figure}[t]
  \centering
  \includegraphics[width=0.9\columnwidth]{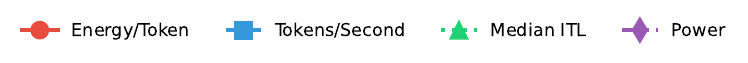}\\
  \begin{subfigure}[b]{0.48\columnwidth}
    \centering
    \includegraphics[width=\columnwidth]{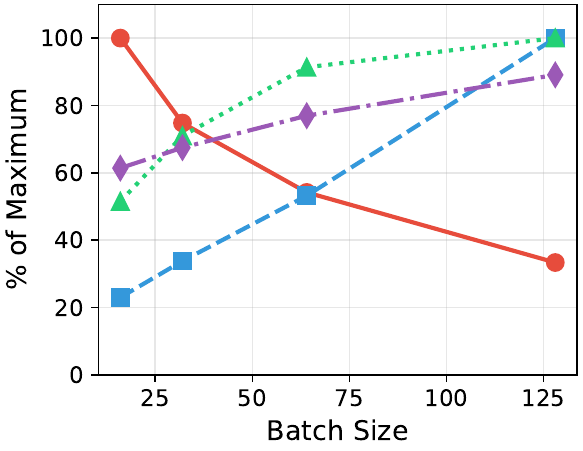}
    \caption{DeepSeek R1}
    \label{fig:batch-size-a}
  \end{subfigure}
  \hfill
  \begin{subfigure}[b]{0.48\columnwidth}
    \centering
    \includegraphics[width=\columnwidth]{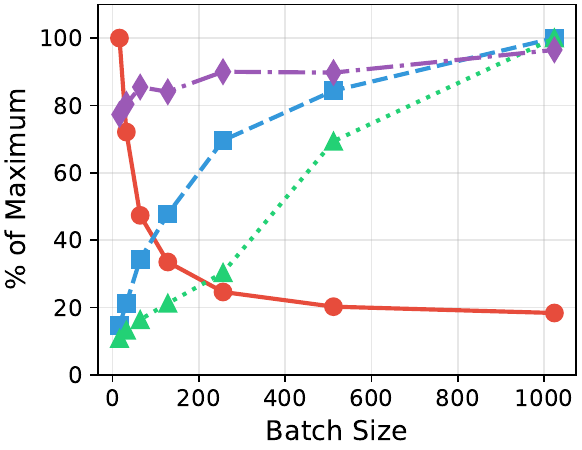}
    \caption{Qwen 3 Coder 30B A3B}
    \label{fig:batch-size-b}
  \end{subfigure}
  \caption{
    Energy per token, throughput, median ITL, and power trends against batch size for (a) DeepSeek R1 (Problem Solving) on 8$\times$ B200 and (b) Qwen 3 Coder 30B A3B (Code Completion) on 1$\times$ B200.
    Metrics normalized to \% of maximum, except power which is normalized to \% of GPU TDP.
  }\label{fig:batch-size}
\end{figure}

Figure~\ref{fig:batch-size} shows the impact of batch size on energy per token, token generation throughput, median Inter-Token Latency (ITL), and GPU power draw.
Computing hardware typically achieves peak energy efficiency when fully utilized (Section~\ref{sec:reasoning} will go deeper into this).
Therefore, as batch size increases, energy per token drops at first, then plateaus as GPU utilization approaches saturation.

However, the energy efficiency gains of increasing batch size are not without tradeoffs.
Latency (median ITL in this analysis) increases with batch size, as there is strictly more work to do for each batch.
Throughput also increases with batch size, but with diminishing returns as GPU utilization reaches saturation.
Finally, power draw increases with batch size, as a larger portion of the GPU's compute and memory circuitry is actively utilized and drawing power.

From energy per token trends, we can see that DeepSeek R1 (Figure~\ref{fig:batch-size-a}) has not saturated GPU utilization even at the largest batch size that fits in memory, whereas Qwen 3 Coder (Figure~\ref{fig:batch-size-b}) approaches saturation around batch size 512.
This explains the two models' throughput trends as well: DeepSeek R1 has a linearly increasing token throughput with batch size as GPU utilization keeps improving, whereas Qwen 3 Coder sees diminishing returns as it approaches saturation.
We can see that these metrics move in tandem rather than in isolation, because they are all heavily coupled with latent factors like GPU utilization.

\begin{takeaway}
Increasing batch size increases latency, power, and throughput, but can unlock 3--5$\times$ energy per token reduction.
\end{takeaway}

\subsection{Model Size and Architecture}\label{sec:model-size}

\begin{figure}[t]
  \centering
  \includegraphics[width=0.76\columnwidth]{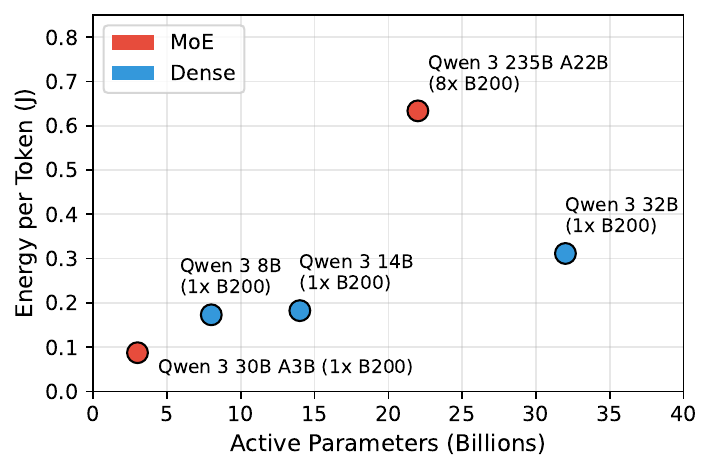}
  \caption{Energy/token by active parameters of Problem Solving models with the minimum-energy configuration on B200 GPUs.}
  \label{fig:model-size}
\end{figure}

With the Mixture-of-Experts (MoE) architecture, the number of active parameters is as important as the total number of parameters in energy consumption.

Figure~\ref{fig:model-size} compares models from the Qwen 3~\cite{qwen3-arxiv25} family on the Problem Solving task using B200 GPUs: two MoE variants (30B A3B and 235B A22B) and three dense variants (8B, 14B, and 32B).
For dense models, energy per token increases with the total number of parameters.
However, when we include MoE models, we see that their energy per token is much lower than what a dense model of similar total number of parameters would consume.
For instance, the energy per token of 30B A3B is 3.56$\times$ lower than that of 32B, despite having a similar total number of parameters.
However, this is not to say that active parameters are now the only factor.
235B A22B consumes more energy than 32B as it needs to use more GPUs to fit all parameters in GPU memory, though it is still far less than what a dense 235B model would consume.

\begin{takeaway}
MoE models consume less energy compared to dense models of similar total number of parameters, making active parameters an important property for energy consumption.
However, total parameters, which affect memory requirements, still play a role.
\end{takeaway}

\subsection{GPU Generation}\label{sec:b200-h100}

\begin{figure}[t]
  \centering
  \includegraphics[width=0.35\columnwidth]{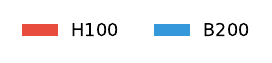}\\[-0.3em]
  \begin{subfigure}[t]{0.48\columnwidth}
    \centering
    \includegraphics[height=3.2cm, valign=t]{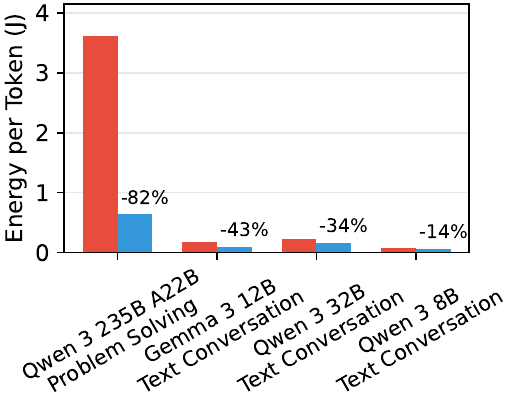}
    \caption{LLM}
    \label{fig:b200-h100-a}
  \end{subfigure}
  \begin{subfigure}[t]{0.48\columnwidth}
    \centering
    \includegraphics[height=3cm, valign=t]{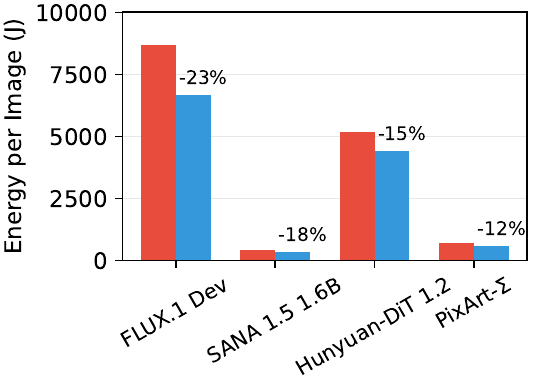}
    \vspace{0.25cm}
    \caption{Text to Image}
    \label{fig:b200-h100-b}
  \end{subfigure}
  \caption{
    B200 vs H100 energy comparison at latency constraints of 100 ms median ITL for LLMs and 30 s generation latency for Text to Image.
    Percentage of B200 energy reduction is annotated.
    See Appendix~\ref{apdx:b200-h100-video} for Text to Video results.
  }\label{fig:b200-h100}
\end{figure}

\looseness=-1
One way to compare GPU models (B200 vs. H100) is to pick the minimum-energy configuration on each GPU at the same latency constraint, as shown in Figure~\ref{fig:b200-h100}.
Energy reduction can vary significantly by model and task.
Sometimes it is significantly better (e.g., 82\% energy per token reduction for Qwen 3 235B A22B Thinking on Problem Solving), other times marginal or even worse, as we will see below.

\looseness=-1
To get a better overall picture, we compare the two GPU models with three different latency constraints: 50/100/250 ms median ITL for LLMs, 10/30/60 s generation latency for Text to Image, 100/500/1000 s for Text to Video.

\paragraph{LLM.}
Across all three median ITL constraints, B200 wins 88\% (63/72) of comparisons with a median 35\% energy reduction (ranging from 53\% more to 82\% less).
A few notable exceptions happen at tight latency constraints.
B200's large VRAM allows fitting large models on fewer GPUs, avoiding inter-GPU communication overhead.
However, at tight latency constraints, using more H100 GPUs with a higher degree of parallelism can be more energy efficient.
For example, at the 50~ms constraint, Qwen 3 30B A3B Thinking uses 53\% less energy on $2\times$ H100 (batch size 128) than on $1\times$ B200 (batch size 64).
Similarly, Qwen 3 235B A22B Instruct FP8 uses 33\% less energy on $8\times$ H100 (batch size 192) than on $2\times$ B200 (batch size 64).
At relaxed constraints ($>$ 50~ms), B200 wins as communication overhead is smaller and higher batch sizes become feasible.
We will look deeper into multi-GPU scaling in Section~\ref{sec:multi-gpu}.

\paragraph{Diffusion.}
For Text to Image, across all three latency constraints, B200 wins 86\% (18/21) of comparisons with a median 15\% energy reduction (ranging from 4\% more to 23\% less).
Text to Video is also similar, with B200 winning 79\% (11/14) of comparisons with a median 4\% energy reduction (ranging from 6\% more to 8\% less).
Cases where H100 wins (e.g., Stable Diffusion 3.5 Medium) are generally when the model is small enough to comfortably fit in one H100 GPU, meaning that it will underutilize a B200.

We performed matched latency constraint comparisons, but we note that B200 would be capable of delivering lower latency than H100 when energy is not a concern due to its higher compute and memory throughput.

\begin{takeaway}
B200 achieves lower energy than H100 in 79--88\% of comparisons at matched latency constraints.
For tight LLM latency constraints, H100 can sometimes consume less energy by using more GPUs with higher parallelism to reduce latency.
For Diffusion, B200 generally wins, unless the model is small.
\end{takeaway}

\subsection{Precision}\label{sec:precision}

\begin{figure}[t]
  \centering
  \includegraphics[width=0.35\columnwidth]{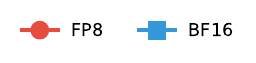}\\
  \begin{subfigure}[b]{0.48\columnwidth}
    \centering
    \includegraphics[width=\columnwidth]{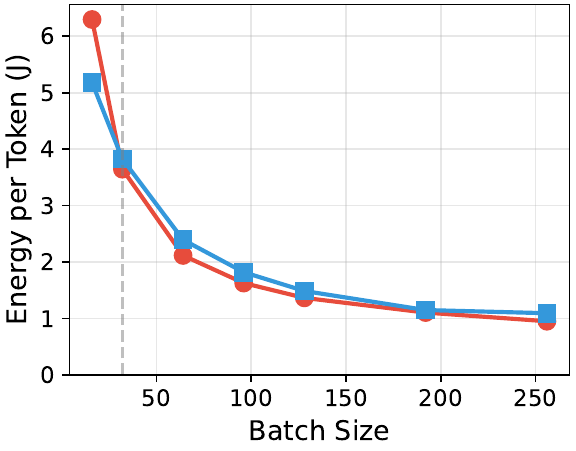}
    \caption{Energy per token}
    \label{fig:precision-a}
  \end{subfigure}
  \hfill
  \begin{subfigure}[b]{0.48\columnwidth}
    \centering
    \includegraphics[width=\columnwidth]{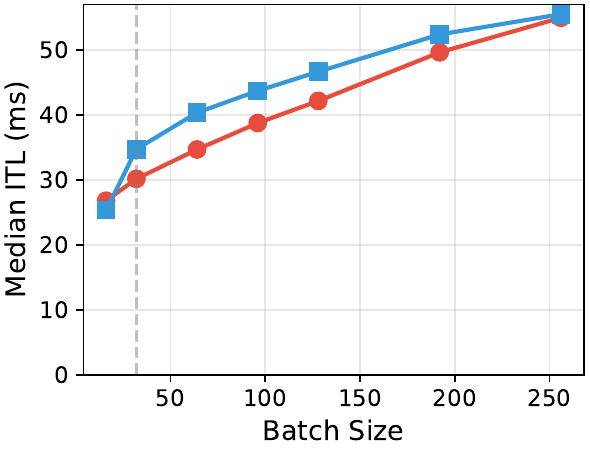}
    \caption{Median ITL}
    \label{fig:precision-b}
  \end{subfigure}
  \caption{Qwen 3 235B A22B (Text Conversation) on $8\times$ H100. FP8 loses at batch size 8--16, then wins at batch sizes from 32. The dashed vertical lines mark the crossover point.}
  \label{fig:precision}
\end{figure}

FP8 quantization reduces model memory footprint and allows inference to leverage FP8 Tensor Cores with higher compute throughput.
However, it also adds overhead from extra operations like input/activation quantization, dequantization, and scaling.
We observe this tradeoff playing out differently at different batch sizes.

\paragraph{FP8 wins at larger batch sizes.}
Figure~\ref{fig:precision} shows the energy per token and median ITL of Qwen 3 235B A22B (Text Conversation) on $8\times$ H100 in both BF16 and FP8 across batch sizes.
At smaller batch sizes, FP8 loses on both energy and latency due to (1) the overhead of extra operations (especially those that have not been fused into matrix multiplication), and (2) underutilization of the GPU, which prevents FP8 from leveraging its compute throughput advantage.
If we compare FP8 and BF16 for all other models and tasks, we see a similar trend:

\begin{center}
\footnotesize
\vspace{-0.5em}
\textbf{Energy}\\[0.3em]
\begin{tabular}{@{}rrrr@{}}
  \toprule
  \textbf{Batch size} & FP8 wins & Range & Median \\
  \midrule
  8--16 & 0/7 & +13 to +56\% & +30\% \\
  17--64 & 6/13 & $-$12 to +32\% & +1\% \\
  65--256 & 11/12 & $-$29 to 0\% & $-$11\% \\
  \bottomrule
\end{tabular}

\vspace{0.5em}
\textbf{Latency}\\[0.3em]
\begin{tabular}{@{}rrrr@{}}
  \toprule
  \textbf{Batch size} & FP8 wins & Range & Median \\
  \midrule
  8--16 & 1/7 & $-$5 to +26\% & +7\% \\
  17--64 & 11/13 & $-$23 to +12\% & $-$12\% \\
  65--256 & 11/12 & $-$18 to +3\% & $-$11\% \\
  \bottomrule
\end{tabular}
\end{center}

\looseness=-1
At batch size 8--16, FP8 has higher energy (up to 56\% more) and higher latency (up to 26\% slower).
As we grow batch size, we see FP8 starting to win on latency earlier, and then on energy as well.
This is, at least in part, because GPUs are capable of delivering more theoretical FP8 compute throughput than BF16.
Thus, at the same batch size, FP8 underutilizes the GPU more, leading to higher energy consumption until batch size is large enough to saturate the GPU.\@

\paragraph{Qwen 3 Coder 480B A35B.}
\looseness=-1
This model is an exception; due to a limitation in vLLM at the time of benchmarking, the FP8 model had to run attention with data parallelism, while BF16 could run attention with tensor parallelism~\cite{vllm-recipes-qwen3-coder}.\footnote{Standard parallelization methods used for LLMs: MoE models use expert parallelism with attention tensor parallelism; dense models use Tensor Parallelism for both MLP and attention.}
This made FP8 consistently consume more time and energy across all batch sizes.
Attention data parallelism incurs load imbalance between GPUs that are assigned very different sequence lengths (e.g., some running long prefills whereas others run decode).
Since the straggler GPU bottlenecks the entire batch, this can lead to significant latency overhead.
Furthermore, the non-straggler GPUs do nothing and waste static power (\S\ref{sec:static-power}) waiting for the straggler, leading to even higher energy consumption as well.

\begin{takeaway}
At smaller batch sizes (8--16), FP8 can consume more time and/or energy than BF16.
FP8 gains start to appear at larger batch sizes.
\end{takeaway}

\subsection{Multi-GPU Scaling}\label{sec:multi-gpu}

We can execute the same model on different numbers of GPUs, which affects both latency and energy consumption.

\begin{figure}[t]
  \centering
  \includegraphics[width=0.35\columnwidth]{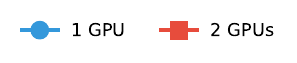}\\
  \begin{subfigure}[b]{0.48\columnwidth}
    \centering
    \includegraphics[width=\columnwidth]{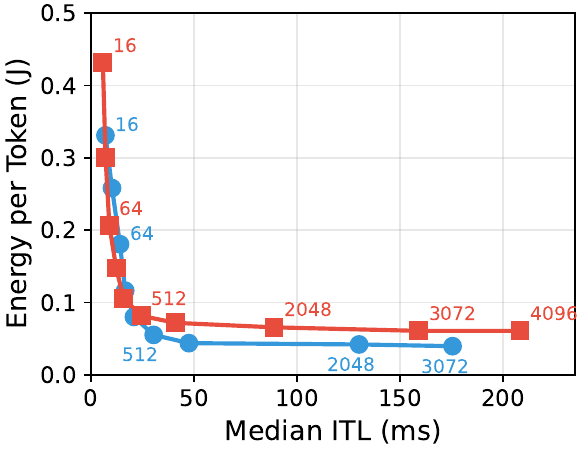}
    \caption{GPT OSS 120B on B200}
    \label{fig:multi-gpu-a}
  \end{subfigure}
  \hfill
  \begin{subfigure}[b]{0.48\columnwidth}
    \centering
    \includegraphics[width=\columnwidth]{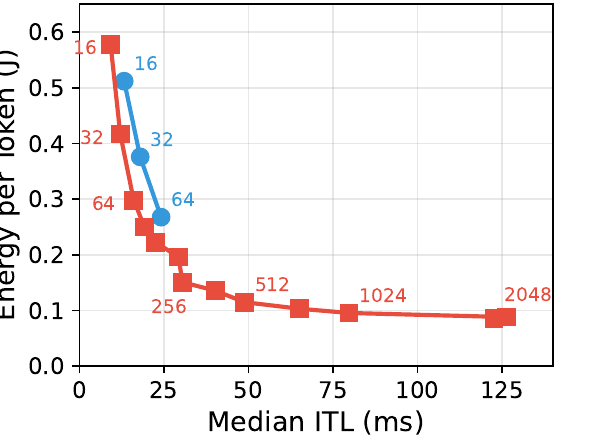}
    \caption{GPT OSS 120B on H100}
    \label{fig:multi-gpu-b}
  \end{subfigure}
  \caption{Time--energy tradeoffs of GPT-OSS 120B (Problem Solving). In both cases, scaling from 1 GPU to 2 GPUs at fixed batch size trades energy for time. In (b), 1 GPU is limited to batch size 64, while 2 GPUs unlock batch size 2,048 with less energy.}
  \label{fig:multi-gpu}
\end{figure}

Figure~\ref{fig:multi-gpu} shows GPT OSS 120B on B200 and H100 with 1 and 2 GPUs.
The plots are time--energy tradeoff curves, which are useful in comparing different configurations:
\begin{denseitemize}
  \item The right-end of each curve represents the minimum-energy configuration for that GPU model and count.
  \item A vertical line at one's target latency finds minimum-energy configurations that meet the latency constraint.
  \item Jumping between curves following points with the same batch size shows the effect of GPU model and count.
\end{denseitemize}

\paragraph{At the same batch size, more GPUs trade energy for latency.}
In general, increasing parallelism with more GPUs reduces latency but also increases energy at the same batch size because (1) latency does not decrease linearly due to communication overhead, and (2) less compute per GPU can lead to lower GPU utilization.
Across B200 configurations, adding GPUs at the same batch size \emph{always} increases energy per token and reduces latency in 81\% of cases.
Similarly, across H100 configurations, energy increases in 93\% of the cases and latency \emph{always} decreases.

\paragraph{Memory capacity-bound cases unlock energy savings with more GPUs.}
On top of the above, in cases where adding more GPUs \emph{enables} larger batch sizes due to increased aggregate memory capacity, we can see energy reductions.
For GPT OSS 120B on 1$\times$ B200 with a 180~GB VRAM, the model already fits at high batch sizes on 1 GPU (batch size 3,072), so 2 GPUs only add overhead without enabling lower energy.
On 1$\times$ H100 with an 80~GB VRAM, however, the server is limited to batch size 64, while 2 GPUs unlock batch size 2,048 and achieve 68\% lower minimum energy.
Thus, the model's total parameter memory footprint relative to the GPU's memory capacity is an important factor for whether multi-GPU scaling can reduce energy.

\begin{figure}[t]
  \centering
  \includegraphics[width=0.85\columnwidth]{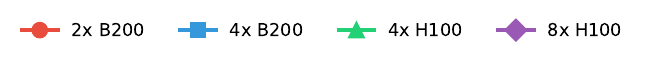}\\[-0.3em]
  \includegraphics[width=0.66\columnwidth]{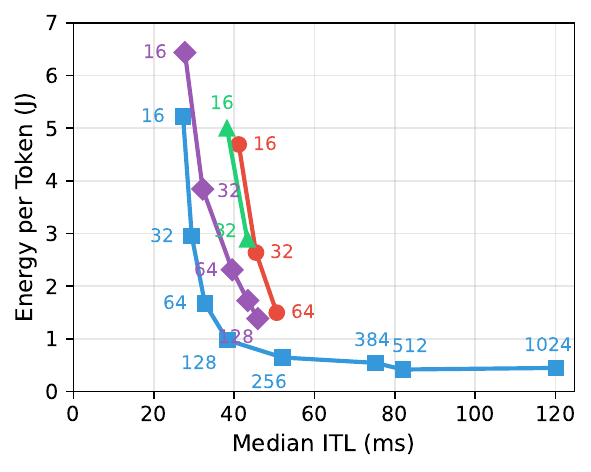}
  \caption{Time--energy tradeoff for Qwen 3 235B A22B Thinking FP8 on Problem Solving across B200 and H100 with different GPU counts. Each point is annotated with its batch size.}
  \label{fig:235b-tradeoff}
\end{figure}

\paragraph{Case study.}
As an extra case study, it is interesting to examine Qwen 3 235B A22B Thinking FP8 on Problem Solving with time--energy tradeoff frontiers for four sets of configurations (2$\times$ and 4$\times$ B200, 2$\times$ and 8$\times$ H100).
As shown in Figure~\ref{fig:235b-tradeoff}, the 4$\times$ B200 curve (blue) Pareto-dominates, and also achieves the lowest possible energy ($\sim$0.4 J/token) by unlocking large batch sizes.
2$\times$ B200 (red) consumes less energy per token compared to 4$\times$ B200 (blue) at the same batch size at the cost of higher latency (as expected), and fails to scale to large batch sizes due to limited memory capacity.
The two H100 configurations (purple and green) are right in the middle of the B200 curves; despite being a whole generation older, H100 is still competitive!

\begin{takeaway}
At the same batch size, more GPUs typically reduce latency but increase energy.
When adding GPUs enables larger batch sizes, energy can be reduced, but only if serving was previously limited by memory capacity.
H100 can still be competitive with B200 in terms of energy, especially when latency constraints are tight.
\end{takeaway}

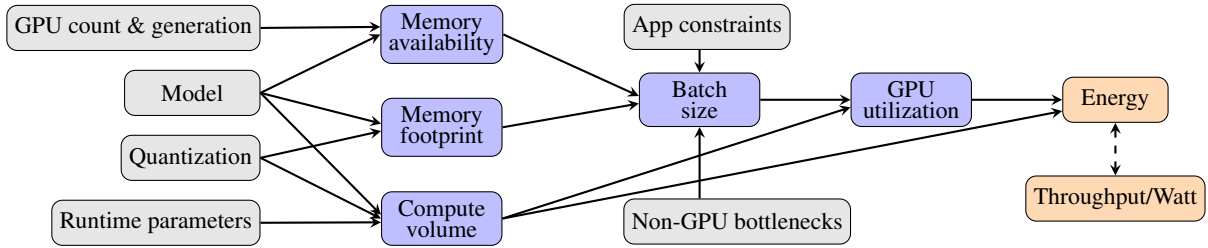
\begin{figure*}[t]
  \centering
  \begin{tikzpicture}[
    node distance=0.4cm and 0.6cm,
    factor/.style={draw, rounded corners, fill=gray!20, font=\footnotesize,
                   minimum height=0.6cm, minimum width=1.8cm, align=center, inner sep=3pt},
    intermediate/.style={draw, rounded corners, fill=blue!25, font=\footnotesize,
                         minimum height=0.6cm, minimum width=1.6cm, align=center, inner sep=3pt},
    output/.style={draw, rounded corners, fill=orange!30, font=\footnotesize,
                   minimum height=0.6cm, minimum width=1.4cm, align=center, inner sep=3pt},
    arrow/.style={->, >=stealth, thick},
    eq/.style={<->, >=stealth, thick, dashed},
  ]

    \node[factor] (hardware) {GPU count \& generation};
    \node[factor, below=0.25cm of hardware.south east, anchor=north east] (modelsize) {Model};
    \node[factor, below=0.25cm of modelsize.south east, anchor=north east] (precision) {Quantization};
    \node[factor, below=0.25cm of precision.south east, anchor=north east] (runtime) {Runtime parameters};

    \node[intermediate, right=4.1cm of hardware.north, anchor=north] (memavail) {Memory\\[-2pt]availability};
    \node[intermediate, below=0.45cm of memavail] (memusage) {Memory\\[-2pt]footprint};
    \node[intermediate, right=3.8cm of runtime.south, anchor=south] (computevol) {Compute\\[-2pt]volume};

    \node[factor, right=1.6cm of memavail, yshift=0.1cm] (appconstraints) {App constraints};
    \node[intermediate, below=0.3cm of appconstraints, xshift=-0.1cm] (batchsize) {Batch\\[-2pt]size};
    \node[factor, below=0.95cm of batchsize, xshift=0.5cm] (cpubottleneck) {Non-GPU bottlenecks};

    \node[intermediate, right=1.2cm of batchsize] (gpuutil) {GPU\\[-2pt]utilization};

    \node[output, right=1.2cm of gpuutil] (energy) {Energy};
    \node[output, below=0.7cm of energy] (tputwatt) {Throughput/Watt};

    \draw[arrow] (hardware.east) -- ([yshift=0.1cm]memavail.west);
    \draw[arrow] (modelsize.east) -- (memavail.west);

    \draw[arrow] (modelsize.east) -- ([yshift=0.05cm]memusage.west);
    \draw[arrow] (precision.east) -- ([yshift=-0.05cm]memusage.west);

    \draw[arrow] (modelsize.east) -- ([yshift=0.05cm]computevol.west);
    \draw[arrow] (precision.east) -- (computevol.west);
    \draw[arrow] (runtime.east) -- ([yshift=-0.05cm]computevol.west);

    \draw[arrow] (memavail.east) -- ([yshift=0.05cm]batchsize.west);
    \draw[arrow] (memusage.east) -- ([yshift=-0.05cm]batchsize.west);
    \draw[arrow] ([xshift=-0.1cm]appconstraints.south) -- (batchsize.north);
    \draw[arrow] ([xshift=-0.5cm]cpubottleneck.north) -- (batchsize.south);

    \draw[arrow] (batchsize.east) -- (gpuutil.west);
    \draw[arrow] (computevol.east) -- ([yshift=-0.1cm]gpuutil.west);

    \draw[arrow] (gpuutil.east) -- (energy.west);
    \draw[arrow] (computevol.east) -- ([yshift=-0.15cm]energy.west);

    \draw[eq] (energy) -- (tputwatt);

  \end{tikzpicture}
  \caption{A framework for reasoning about inference energy consumption in our analysis.
    \tikz[baseline=-0.5ex]{\node[draw, rounded corners, fill=gray!20, font=\scriptsize, inner sep=2pt] {Gray boxes};} are properties and knobs,
    \tikz[baseline=-0.5ex]{\node[draw, rounded corners, fill=blue!25, font=\scriptsize, inner sep=2pt] {blue boxes};} are latent factors, and
    \tikz[baseline=-0.5ex]{\node[draw, rounded corners, fill=orange!30, font=\scriptsize, inner sep=2pt] {orange boxes};} are the end metrics we observe from measurements and would like to understand and explain.}
  \label{fig:energy-factors}
\end{figure*}

\section{Reasoning about Energy Consumption}\label{sec:reasoning}

In the previous sections, we presented empirical observations on energy consumption, but how can we act on them?
In this section, we outline core mechanisms that govern energy consumption, with the goal of providing tools to \emph{explain and reason about} energy consumption.

\subsection{Model, Runtime, and Hardware Factors}\label{sec:core-mechanisms}

\looseness=-1
Many factors across the whole system (hardware, software, and algorithm) affect energy consumption.
Some of the key mechanisms are powerful but still straightforward.
For instance, more computation generally means more energy consumption.
As we have seen, diffusion models' energy increases with more denoising steps and higher output resolution (\S\ref{sec:diffusion}), MoE models activate fewer parameters per token than dense models (\S\ref{sec:model-size}), and FP8 reduces circuit activity via lower-precision arithmetic (\S\ref{sec:precision}).
These are examples of choices at the runtime- and model-level directly affecting the amount of computation, and thus energy consumption.

Another instance is hardware efficiency improvements over generations.
Newer architectures typically deliver more operations per joule via various microarchitectural improvements and technology node shrinks.
We have indeed seen that B200 generally consumes less energy than H100 (\S\ref{sec:b200-h100}).

\subsection{Static Power Wastage}\label{sec:static-power}

The power consumption of computing hardware, including GPUs, has two components: \emph{static power} (consumed regardless of activity at all times) and \emph{dynamic power} (reflects compute and memory activity).
Let us consider a case where we executed some computation on a GPU, and only 60\% of the GPU's compute units were utilized over the entire execution time.
Here, the GPU will consume static power for the entire execution time, regardless of how well the GPU is utilized.
Thus, 40\% of the time the GPU is consuming static power while making little progress, effectively wasting energy.
This is how low utilization increases static power wastage and thus energy consumption for the same amount of work.

However, one of the most critical factors in GPU utilization is, in fact, not the GPU, but the rest of the system.
That is, we want the GPU to be the sole bottleneck, not other system components.
When CPU processing, network communication, disk I/O, or other parts of the system block GPU progress, the GPU does not have enough work to saturate itself or is even idle, wasting static power.
Multimodal LLMs (\S\ref{sec:mllm}) were a prime example: CPU-side vision preprocessing became a bottleneck that limited batch size, leaving the GPU underutilized despite having capacity for more concurrent requests.
The result was higher energy per token---not because of the GPU, but because of the surrounding system.

Another important factor is arithmetic intensity, i.e., the ratio of compute operations to the amount of memory movement.
When arithmetic intensity is low, the GPU may be waiting on memory fetches more often than performing computations, leading to lower GPU utilization and higher static power wastage.
We observed this for precision (\S\ref{sec:precision}), where FP8 computations require extra operations that are not as arithmetically intensive as matrix multiplications.
Thus, on smaller batch sizes, both FP8 extra operations and the smaller matrix multiplications had lower arithmetic intensity, leading to lower GPU utilization and offsetting savings from lower-precision arithmetic.

This has interactions with earlier factors as well.
For instance, upgrading to a newer hardware generation expecting better energy efficiency may not yield the expected benefits, or even worsen, if bottlenecks in the rest of the system were preventing the GPU from being fully utilized.

\subsection{Time--Energy Tradeoff Frontier}\label{sec:tradeoff-frontier}

There are many cases where there is a time--energy tradeoff frontier for the same amount of work (e.g., Figure~\ref{fig:235b-tradeoff}).
When the GPU ideally \emph{is} the bottleneck, largely ruling out static power wastage (\S\ref{sec:static-power}), we can navigate the time--energy tradeoff frontier through configuration choices.\footnote{When the GPU is being underutilized, a proper time--energy \emph{tradeoff} frontier may not exist, as both time and energy can be reduced by improving GPU utilization.}
In our analysis, the factors that govern this frontier are:
\begin{denseitemize}
  \item \textbf{Batch size:} This is the primary knob that \emph{shapes} and \emph{navigates} the time--energy frontier.
  \item \textbf{Memory capacity:} Larger batches consume more memory. When GPU memory is saturated, we hit a ceiling, like we have seen for reasoning models in Section~\ref{sec:llm}. In other words, memory capacity \emph{bookends} the frontier.
  \item \textbf{Application constraints:} Applications may come with latency deadlines or energy budgets. Larger batches increase per-request latency and reduce energy per work. Application-level latency and/or energy budgets allow us to \emph{select} a point on the frontier.
\end{denseitemize}

Batch size does not have to be the only knob that shapes the time--energy tradeoff frontier.
For instance, the number of GPUs (\S\ref{sec:multi-gpu}) can be effective, where adding GPUs increases aggregate memory capacity and also enables larger batch sizes that were not previously possible.
While not explored in this paper, GPU power limit~\cite{zeus-nsdi23} and core frequency~\cite{perseus-sosp24,dynamollm-hpca25,kareus-arxiv26} are also core knobs that shape the frontier.

\subsection{Extending to AI Datacenters}\label{sec:power}

Our analysis so far has focused on energy consumption, but power is also an important metric to consider.
Indeed, many AI datacenters today are \emph{power-constrained}~\cite{cbre2025,bloombergnef25,inferencemax-blog25}.
Power availability caps the datacenter's power budget---either from the electricity grid (where drawing too much may not be approved or may cause reliability issues~\cite{ai-grid-impact-arxiv25}) or from on-site generation like natural gas and batteries (which take \emph{years} to build~\cite{eia-utility-data}).

With power becoming the bottleneck resource, \emph{throughput per watt} (e.g., tokens per second per watt, images per second per watt) is a critical metric for AI datacenter operators.
For instance, tokens per second per watt can tell the operator how many average ChatGPT users the datacenter can serve within its power budget.
\begin{equation*}
  \frac{\text{Throughput}}{\text{Power}} = \frac{\text{Work}/\text{Time}}{\text{Energy}/\text{Time}} = \frac{\text{Work}}{\text{Energy}}
\end{equation*}
Throughput per watt is essentially the inverse of energy consumption per fixed work (e.g., energy per token, energy per image).
Thus, optimizing energy consumption for the given work improves throughput per watt, closing the reasoning loop.
Appendix~\ref{apdx:power} provides a quantitative evaluation.

\subsection{Putting Everything Together}\label{sec:putting-together}

Figure~\ref{fig:energy-factors} summarizes the structure of reasoning we have developed.
\tikz[baseline=-0.55ex]{\node[draw, rounded corners, fill=gray!20, font=\footnotesize, inner sep=2pt]{Gray boxes};} are \emph{properties} and \emph{low-level knobs} of the algorithm, software, and hardware.
\tikz[baseline=-0.73ex]{\node[draw, rounded corners, fill=blue!25, font=\footnotesize, inner sep=2pt]{Blue boxes};} represent \emph{latent} variables that mediate between configurations and outcomes.
\tikz[baseline=-0.55ex]{\node[draw, rounded corners, fill=orange!30, font=\footnotesize, inner sep=2pt]{Orange boxes};} show the end metrics we measure from benchmarks and ultimately want to understand.

Causal structures like this show how different factors interact and propagate to affect the end metric and provide a framework for explaining empirical observations.
When we observe unexpected energy behavior, we can trace through these factors to identify the root cause---whether it is memory constraints limiting batch size, CPU bottlenecks causing GPU underutilization, or compute volume increasing due to model choices.
This also enables reasoning about optimization opportunities and hypothesizing about how they will affect energy consumption.

\section{Related Work}\label{sec:related}

We build on top of the ML.ENERGY Benchmark~\cite{mlenergy-neurips25}, which provides facilities for measuring the energy of inference under realistic conditions.
MLPerf Power~\cite{mlperfpower-hpca25} is an industry-backed benchmark for power, but it focuses on a small set of most important models and tasks, unable to capture the diversity of factors affecting energy consumption.
The Hugging Face AI Energy Score~\cite{ai-energy-score} fixes batch size to 1, and GPU underutilization (\S\ref{sec:static-power}) would inflate energy numbers and render them unrepresentative.
Google disclosed the median energy consumption of their AI service~\cite{google-ai-energy-arxiv25} with the most comprehensive scope that includes idle machine overhead, but measurements are on internal systems, limiting generalizability.
Most importantly, existing works do not go deep into mapping the latent factors and causal structures that manifest as energy differences.

\section{Conclusion}\label{sec:conclusion}

As energy becomes a critical bottleneck in scaling AI infrastructure, understanding inference energy consumption---not just measuring it---is essential for systematic optimization.
With observations from a large-scale up-to-date empirical study, we mapped out the factors and their relationship with energy consumption: time and energy are governed by latent factors (e.g., compute, memory, utilization, and application constraints) that mediate between configuration choices and end metrics.
This framework enables moving beyond black-box observations and tracing how model, system, and application factors propagate to energy consumption.

\section*{Impact Statement}
This paper presents work whose goal is to advance the field of machine learning and systems support for machine learning. There are many potential societal consequences of our work, none of which we feel must be specifically highlighted here.

\bibliography{ref}
\bibliographystyle{icml2026}

\clearpage
\appendix

\section{List of Tasks and Models}\label{apdx:tasks-and-models}

\begin{table*}[t]
\centering
\caption{Benchmark tasks and their request datasets.}
\label{tab:apdx-tasks}
\begin{tabular}{lp{4.2cm}l}
\toprule
\textbf{Task} & \textbf{Description} & \textbf{Request Dataset} \\
\midrule
\multicolumn{3}{l}{\textit{Large Language Models (LLM)}} \\
Problem Solving & Graduate-level science questions requiring deep reasoning & GPQA Diamond~\cite{gpqa-colm24} \\
Text Conversation & General conversational benchmark & Arena Human Preference~\cite{lmarena-dataset} \\
Code Completion & Fill-in-the-middle code completion & Sourcegraph Context-Aware FIM~\cite{sourcegraph-fim} \\
\midrule
\multicolumn{3}{l}{\textit{Multimodal Large Language Models (MLLM)}} \\
Image Chat & Multimodal image understanding and chat & VisionArena Chat~\cite{visionarena-dataset} \\
Video Chat & Multimodal video understanding and chat & LLaVA-Video-178K~\cite{llavavideo-tmlr25} \\
\midrule
\multicolumn{3}{l}{\textit{Diffusion Models}} \\
Text to Image & Generate images from text prompts & Open Image Preferences~\cite{open-image-preferences} \\
Text to Video & Generate videos from text prompts & EvalCrafter~\cite{evalcrafter-cvpr24} \\
\bottomrule
\end{tabular}
\end{table*}

\begin{table*}[t]
\centering
\caption{Models benchmarked for each task. Models are BF16 unless noted as FP8.}
\label{tab:apdx-models}
\small
\begin{tabular}{lp{14cm}}
\toprule
\textbf{Task} & \textbf{Models} \\
\midrule
\multicolumn{2}{l}{\textit{Large Language Models (LLM)}} \\
Problem Solving & Qwen 3 8B, 14B, 32B, 30B A3B Thinking, 235B A22B Thinking, 235B A22B Thinking FP8~\cite{qwen3-arxiv25}; DeepSeek R1~\cite{deepseek-r1-arxiv25}, V3.1~\cite{deepseekv31-hf}; NVIDIA Nemotron Nano 9B V2, 12B V2~\cite{nemotron-nano-arxiv25}; GPT OSS 20B, 120B~\cite{gpt-oss-arxiv25} \\
\addlinespace
Text Conversation & Qwen 3 8B, 14B, 32B, 30B A3B Instruct, 235B A22B Instruct, 235B A22B Instruct FP8~\cite{qwen3-arxiv25}; Gemma 3 12B, 27B~\cite{gemma3-arxiv25}; Llama 3.1 8B, 70B, 405B, 405B FP8, Llama 3.3 70B~\cite{llama3-arxiv24}; Llama 4 Scout 17B 16E, Maverick 17B 128E FP8~\cite{llama4-blog}; DeepSeek V3.1~\cite{deepseekv31-hf}; NVIDIA Nemotron Nano 9B V2, 12B V2~\cite{nemotron-nano-arxiv25} \\
\addlinespace
Code Completion & Qwen 3 Coder 30B A3B, 480B A35B, 480B A35B FP8~\cite{qwen3-arxiv25} \\
\midrule
\multicolumn{2}{l}{\textit{Multimodal Large Language Models (MLLM)}} \\
Image Chat & Qwen 3 VL 8B, 32B, 30B A3B, 235B A22B, 235B A22B FP8, Qwen 3 Omni 30B A3B~\cite{qwen3-omni-arxiv25}; Gemma 3 12B, 27B~\cite{gemma3-arxiv25}; Llama 4 Scout 17B 16E, Maverick 17B 128E FP8~\cite{llama4-blog}; NVIDIA Nemotron Nano 12B V2 VL~\cite{nemotron-nano-v2-vl-arxiv25} \\
\addlinespace
Video Chat & Qwen 3 VL 8B, 32B, 30B A3B, 235B A22B, 235B A22B FP8~\cite{qwen3-vl-arxiv25}, Qwen 3 Omni 30B A3B~\cite{qwen3-omni-arxiv25} \\
\midrule
\multicolumn{2}{l}{\textit{Diffusion Models}} \\
Text to Image & Stable Diffusion 3.5 Medium, Large~\cite{sd3-icml24}; FLUX.1 Dev~\cite{flux1dev-2024}; PixArt-$\Sigma$~\cite{pixart-sigma-eccv24}; Hunyuan-DiT 1.2~\cite{hunyuandit-arxiv24}; SANA 1.5 1.6B, 4.8B~\cite{sana-arxiv24} \\
\addlinespace
Text to Video & CogVideoX 2B, 1.5 5B~\cite{cogvideox-iclr25}; Wan 2.1 1.3B, 14B~\cite{wan-arxiv25}; HunyuanVideo~\cite{hunyuanvideo-arxiv24} \\
\bottomrule
\end{tabular}
\end{table*}

Table~\ref{tab:apdx-tasks} lists the benchmark tasks and their corresponding request datasets used in the measurement studies in Sections~\ref{sec:energy-by-arch} and~\ref{sec:deeper-dive}, and Table~\ref{tab:apdx-models} lists the models.

\section{B200 vs H100 for Text to Video}\label{apdx:b200-h100-video}

\begin{figure}[t]
  \centering
  \includegraphics[width=0.35\columnwidth]{figures/section2-3-b200-vs-h100-legend.pdf}\\[0.3em]
  \includegraphics[width=0.5\columnwidth]{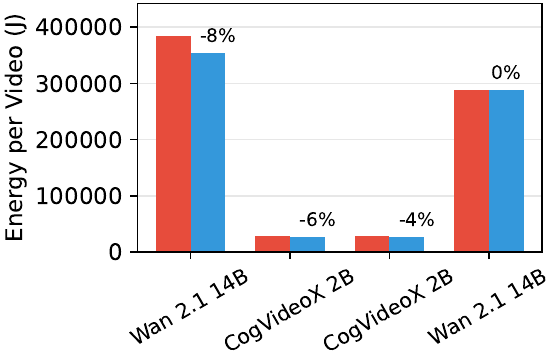}
  \caption{B200 vs H100 energy comparison for Text to Video at a latency constraint of 500~s generation latency. Percentage of B200 energy reduction is annotated.}
  \label{fig:b200-h100-video}
\end{figure}

Figure~\ref{fig:b200-h100-video} shows the B200 vs H100 energy comparison for Text to Video diffusion models at a latency constraint of 500~s generation latency.
The annotations indicate the energy reduction percentage of B200 relative to H100.

\section{Throughput per Watt}\label{apdx:power}

\begin{figure}[t]
  \centering
  \includegraphics[width=\columnwidth]{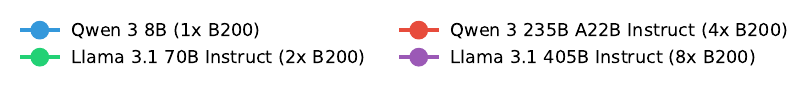}\\[0.5em]
  \begin{subfigure}[b]{0.48\columnwidth}
    \centering
    \includegraphics[width=\columnwidth]{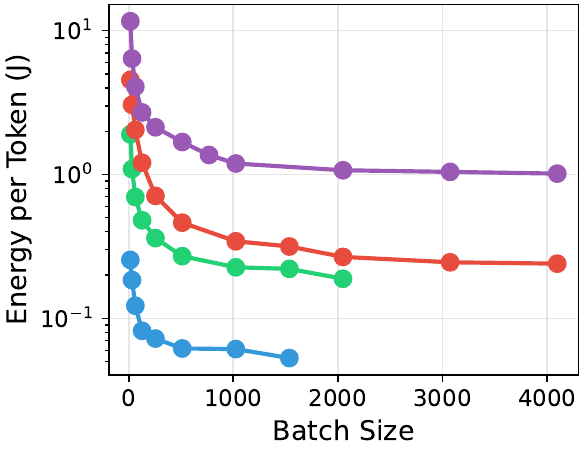}
    \caption{Energy per token}
    \label{fig:power-a}
  \end{subfigure}
  \hfill
  \begin{subfigure}[b]{0.48\columnwidth}
    \centering
    \includegraphics[width=\columnwidth]{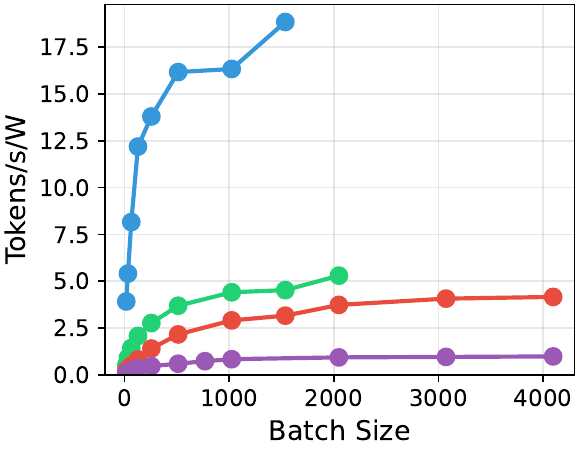}
    \caption{Throughput per watt}
    \label{fig:power-b}
  \end{subfigure}
  \caption{Energy and throughput/watt for four models on B200 with varying batch size. Note the log scale Y axis in (a).}
  \label{fig:power}
\end{figure}

Figure~\ref{fig:power} shows energy per token and throughput per watt for four LLMs on B200 across batch sizes.
As batch size increases, energy per token decreases (better efficiency) and throughput per watt increases (more work done per unit power).
Since throughput per watt is the inverse of energy per work, optimizing one directly optimizes the other.

\end{document}